%% file: main.tex
\documentclass[letterpaper, 10 pt, conference]{ieeeconf}  %

\IEEEoverridecommandlockouts                              %

\usepackage{amsmath,amssymb,amsfonts}
\usepackage{xcolor}
\usepackage{cite}
\usepackage{pifont} %
\usepackage{graphicx}
\usepackage{bigstrut}
\usepackage{booktabs}  %
\usepackage{colortbl}  %
\usepackage{multirow}  %
\usepackage{siunitx}  %
\usepackage[font={footnotesize}]{caption}
\usepackage[colorlinks]{hyperref}
\usepackage{rotating}  %

\definecolor{darkgreen}{RGB}{0,127,0}
\definecolor{darkred}{RGB}{200,0,0}

\newcommand{\link}[1]{{\color{blue}\href{#1}{#1}}}
\newcommand{\xmark}{\text{\ding{55}}}  %
\def\greencheckmark{\textcolor{darkgreen}{\checkmark}}
\def\redxmark{\textcolor{darkred}{\xmark}}
\newcommand{\tH}[1]{\textbf{#1}}

\newenvironment{myitem}{\begin{list}{$\bullet$}
{\setlength{\itemsep}{-0pt}
\setlength{\topsep}{0pt}
\setlength{\labelwidth}{5pt}
\setlength{\leftmargin}{10pt}
\setlength{\parsep}{-0pt}
\setlength{\itemsep}{0pt}
\setlength{\partopsep}{0pt}}}%
{\end{list}}

\title{\LARGE \bf
HANDAL:  A Dataset of Real-World Manipulable Object Categories \\ with Pose Annotations, Affordances, and Reconstructions
}

\author{Andrew Guo, Bowen Wen, Jianhe Yuan, Jonathan Tremblay, Stephen Tyree, Jeffrey Smith, Stan Birchfield \\ NVIDIA}

\begin{document}

\twocolumn[{
\renewcommand\twocolumn[1][]{#1}%
\maketitle

\begin{center}
    \centering
     \includegraphics[width = 0.93\textwidth]{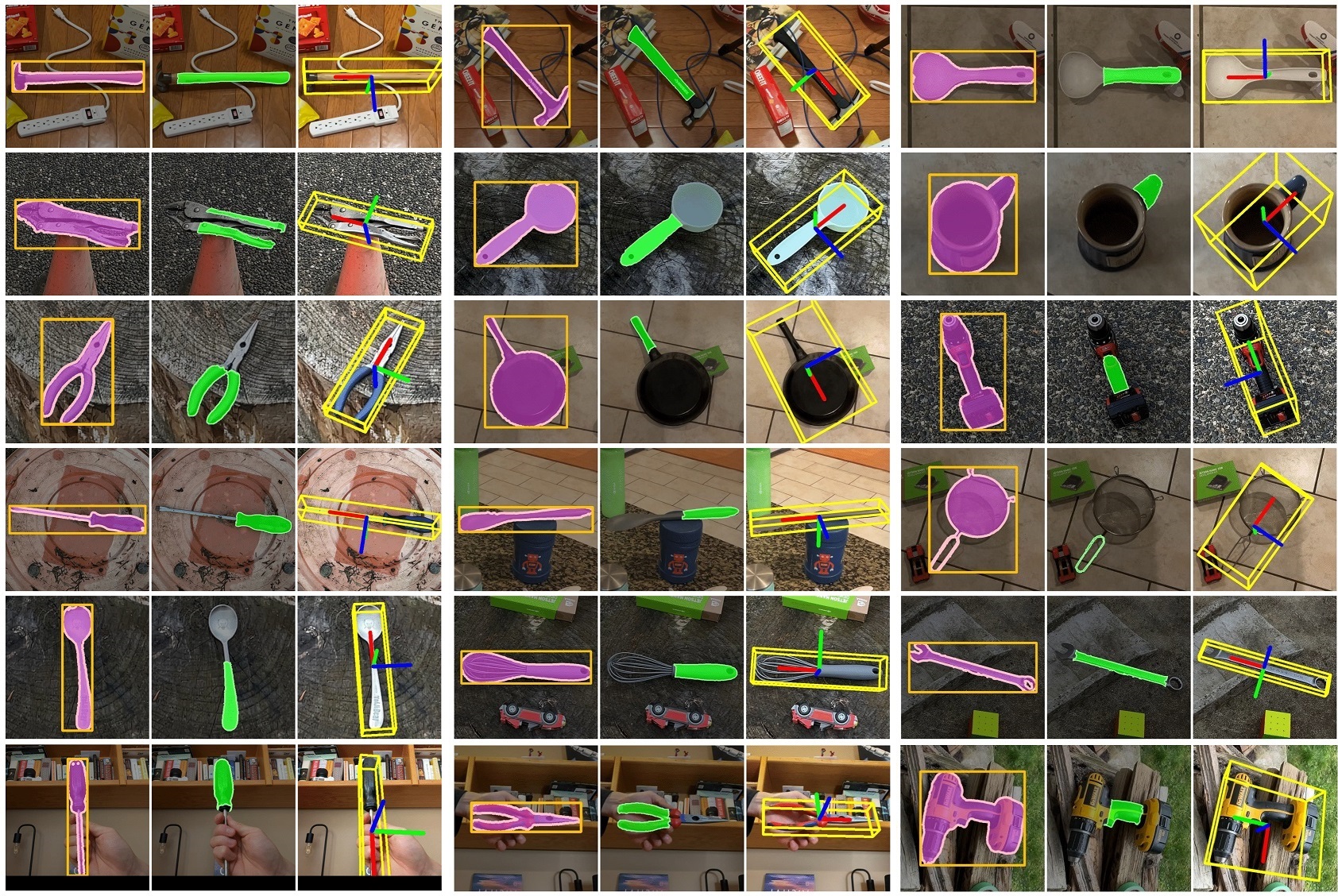}
    \captionof{figure}{Gallery of our HANDAL dataset which covers a variety of object categories that are friendly for robotic manipulation and functional grasping; it includes both static and dynamic scenes. Each subfigure shows precise ground-truth annotations of an image:   2D bounding box (orange), pixel-wise object segmentation (purple) obtained by projecting the reconstructed 3D model, pixel-wise projection of 3D handle affordance segmentation (green), and 6-DoF category-level pose+scale. Image contrast is adjusted for visualization purposes.}\label{fig:intro}
\end{center}%
}]

\thispagestyle{empty}
\pagestyle{empty}

\begin{abstract}

We present the HANDAL dataset for category-level object pose estimation and affordance prediction.  Unlike previous datasets, ours is focused on robotics-ready manipulable objects that are of the proper size and shape for functional grasping by robot manipulators, such as pliers, utensils, and screwdrivers.
Our annotation process is streamlined, requiring only a single off-the-shelf camera and semi-automated processing, allowing us to produce high-quality 3D annotations without crowd-sourcing.
The dataset consists of 308k annotated image frames from 2.2k videos of 212 real-world objects in 17 categories.
We focus on hardware and kitchen tool objects to facilitate research in practical scenarios in which a robot manipulator needs to interact with the environment beyond simple pushing or indiscriminate grasping.
We outline the usefulness of our dataset for 6-DoF category-level pose+scale estimation and related tasks. 
We also provide 3D reconstructed meshes of all objects, and we outline some of the bottlenecks to be addressed for democratizing the collection of datasets like this one. 
Project website: \link{https://nvlabs.github.io/HANDAL/}

\end{abstract}

\section{INTRODUCTION}

If robots are to move beyond simple pushing and indiscriminate top-down grasping, they must be equipped with detailed awareness of their 3D surroundings~\cite{tang2023rgb}.
To this end, high-quality 3D datasets tailored to robotics are needed for training and testing networks.
Compared to the many large-scale 2D computer vision datasets (e.g., ImageNet~\cite{deng2009imagenet}, COCO~\cite{lin2014coco}, and OpenImages~\cite{kuznetsova2020openimages}), existing 3D robotics datasets tend to be rather small in size and scope.

The primary challenge in 3D is the annotation of real images, which requires either depth or multiple images, thus making it difficult to scale up annotation processes.
While the datasets of the BOP challenge~\cite{hodan2018eccv:bop} have filled this gap for instance-level object pose estimation,
the challenge still remains for category-level object pose estimation, as well as for learning functional affordances---such as object handles---for task-oriented grasping and manipulation~\cite{wen2022catgrasp,wen2022you,huang2022parallel,gao2022toward,gao2021minimizing}.

To address these problems, we introduce the HANDAL dataset.\footnote{``\emph{H}ousehold \emph{AN}notated \emph{D}ataset for function\emph{AL} robotic grasping and manipulation''.  The name also fits because all of our objects have handles.}
Using an off-the-shelf camera to collect data, and a semi-automated pipeline for annotating the data in 3D, we have created a rather large labeled dataset without much human effort (once the pipeline was set up).
Fig.~\ref{fig:intro} shows examples of our image frames with precise 3D annotations including segmentation, affordances (handles), 
6-DoF poses, and bounding boxes.
We see this dataset as an important step toward democratizing the creation of labeled 3D datasets.  
To this end, we do not use any specialized hardware for collection, nor any crowdsourcing for annotation. 
Our dataset has the following properties:
\begin{myitem}
\item Short video clips of object instances from 17 categories that are well-suited for robotic functional grasping and manipulation, with 6-DoF object pose+scale annotations for all image frames.  Multiple captures of each instance ensure diversity of backgrounds and lighting.
  \item 3D 
  reconstructions of all objects, along with task-inspired affordance annotations.
  \item Additional videos of the objects being handled by a human to facilitate task-oriented analysis in dynamic scenarios.  These dynamic videos are also equipped with accurate ground-truth annotations.
\end{myitem}
At the end of the paper we include a discussion of remaining bottlenecks for the process of creating a dataset like ours.

\section{Previous Work}

\textbf{Large-object datasets.}
Precursors to the present work include datasets for autonomous driving, such as KITTI 3D object detection~\cite{geiger2013ijrr:kitti} and nuScenes~\cite{caesar2020cvpr:nuscenes}, where the objects to be detected are vehicles, cyclists, and pedestrians.
Similarly, datasets like PASCAL3D+~\cite{xiang2014wacv:pascal}
and SUN RGB-D~\cite{song2015cvpr:sunrgbd} enable 3D object detection of tables, chairs, couches, and other large indoor items.
In both cases the objects are much too large for robotic manipulation, and their upright orientation generally precludes the need for full 6-DoF pose estimation.

\textbf{Category-level object pose datasets.}
The first dataset of real images for small, manipulable objects was NOCS-REAL275~\cite{wang2019normalized}, which consists of a small number of annotated RGBD videos of 
six categories:  bottle, bowl, camera, can, laptop and mug.
Building on this work, Wild6D~\cite{fu2022category} expands the number of videos and images, but with similar categories:  bottle, bowl, camera, laptop, and mug.

More recent datasets~\cite{wang2022phocal,jung2022arx:housecat} address the same problem as ours, but on a smaller scale.
PhoCaL~\cite{wang2022phocal} contains RGBD videos of eight categories:  bottle, box, can, cup, remote, teapot, cutlery, and glassware.  
HouseCat6D~\cite{jung2022arx:housecat} contains RGBD videos of a similar set of ten categories:  bottle, box, can, cup, remote, teapot, cutlery, glass, shoe, and tube.
Similar to our work, the objects in these categories are potentially manipulable by a robot.

\input{tables/comparison.tex}

Objectron~\cite{ahmadyan2021objectron} scales up the number of object instances and videos considerably, but most of the categories are not manipulable.
This dataset is therefore more applicable to computer vision than to robotics.
Similarly, CO3D~\cite{reizenstein2021iccv:co3d} contains a large number of videos of a large number of categories, and most of these are not manipulable.
CO3D is different, however, in that its focus is on category-specific 3D reconstruction and novel-view synthesis, rather than category-level object pose estimation.  As a result, this dataset omits some key qualities that are required for the latter, such as absolute scale and object pose with respect to a canonical coordinate system.

In contrast to these works, our focus is on manipulable objects, particularly those well-suited for robotic functional grasping.
As such, we collect data of relatively small objects with handles, and we annotate their affordances as well as 6-DoF pose+scale.
See Table~\ref{tab:comparison} for a comparison between our 3D dataset and related works.

\textbf{Category-level object pose estimation.}
Category-level object pose estimation is a relatively recent problem that has been gaining attention lately.
A natural extension of instance-level object pose estimation~\cite{hodan2018eccv:bop,calli2015ram:ycb,xiang2018rss:posecnn,tyree2022iros:hope,wen2020robust,wen2020se}, category-level pose estimation~\cite{wang2019normalized,chen2021sgpa,chen2021fs,lee2023tta,wen2021bundletrack} addresses broad classes of objects instead of known sets of object instances, and is therefore potentially more useful for robotics applications.
The category-level datasets mentioned above have enabled recent research in this important area.

In their seminal work, Wang et al.~\cite{wang2019normalized} introduced the normalized object coordinate system (NOCS), which was then used to perform 3D detection and category-level object pose estimation from a single RGBD image.  This method was evaluated on the NOCS-REAL275 dataset.
In follow-up work, CPS~\cite{manhardt2000arx:cpspp}
also evaluates on REAL275, but this method only requires an RGB image at inference time.  The method was trained in a self-supervised manner using unlabeled RGBD images, and it also infers the shape of the instance as a low-dimensional representation.
The method of DualPoseNet~\cite{lin2021iccv:dualposenet} uses a pair of decoders, along with spherical fusion, to solve this problem using a single RGBD image.  The method was also evaluated on NOCS-REAL275.
More recent approaches of SSP-Pose~\cite{zhang2022iros:ssppose}, CenterSnap~\cite{irshad2022icra:centersnap}, iCaps~\cite{deng2022ral:icaps}, and ShAPO~\cite{irshad2022eccv:shapo} all evaluate on NOCS-REAL275.

MobilePose~\cite{hou2020:mobilepose} aims to recover object pose from a single RGB image, evaluating on the Objectron dataset.
CenterPose~\cite{lin2022icra:centerpose} and its tracking extension, CenterPoseTrack~\cite{lin2022icra:centerposetrack}, address the same problem, using a combination of keypoints and heatmaps, connected by a convGRU module.
These two methods were also evaluated on Objectron.

To our knowledge, previous methods have been evaluated on only two datasets:  REAL275 and Objectron.
We aim to extend this line of research by providing the largest dataset yet focused on categories that are amenable to robotic manipulation, along with annotations that facilitate functional grasping.

\section{Dataset overview}

Our goal with this dataset is to support research in perception that will enable robot manipulators to perform real-world tasks beyond simple pushing and pick-and-place.  
As a result, we were motivated to select object categories with functional purposes.
In particular, we focused on categories with a handle to facilitate functional grasps.

\subsection{Object categories}

We have selected 17 categories of graspable, functional objects: claw hammer, fixed-joint pliers, slip-joint pliers, locking pliers, power drill, ratchet, screwdriver, adjustable wrench, combination wrench, ladle, measuring cup, mug, pot/pan, spatula, strainer, utensil, and whisk.  At a high level, these categories generally fall into one of two larger super-categories:
objects that might be found in a toolbox (e.g., hammers)
and objects that might be found in a kitchen (e.g., spatulas).

Because all the objects were designed for human handling, they are of an appropriate size for robot grasping and manipulation, with two caveats.  First, some objects may be too heavy for existing robot manipulators; we envision that 3D printed replicas or plastic versions could be substituted for actual manipulation experiments.
Secondly, since all our objects support functional grasping, that is, grasping for a particular use---as opposed to indiscriminate grasping---some objects may require anthropomorphic robotic hands rather than parallel-jaw grippers.  
We expect that our dataset will open avenues for further research in this area.

The objects are composed of a variety of materials and geometries, including some with reflective surfaces and some with perforated or thin surfaces.
The many challenges introduced by such properties should make our dataset useful for furthering perception research to deal with real-world scenarios.

Some of the objects in our dataset allow for articulation, but we treat the articulated objects as if they were rigid by choosing a default state for each instance before capture.
Nevertheless, we believe that our annotations set the stage for future research in this important area.

\subsection{Comparison with other datasets}
\label{sec:compare}

To put this work in context, our dataset is compared with existing category-level object datasets in Tab.~\ref{tab:comparison}.  From left to right, the table captures the input modality, number of categories, number of manipulable categories, number of object instances, number of videos, number of images/frames, whether object pose is recorded, whether absolute scale is available, whether 3D reconstructions are provided, whether the videos include static (object not moving) and/or dynamic (object moving) scenes, whether videos capture 360$^\circ$ views of the objects, whether objects are partially occluded in some views/scenes, and whether object affordances are annotated.

Our dataset is unique because it contains instance- as well as category-level pose annotations, since we captured multiple videos per instance.  As a result, our data can also be used to train an instance-level pose estimator that operates with respect to a specific 3D object model.
We also have the ability to reconstruct 3D models of the objects and align them across videos, which yields pose annotations that are much more precise than is possible with manual annotation.

Determining whether an object is manipulable is somewhat subjective.  For the table above, we do not include cameras or laptops (from NOCS-REAL275~\cite{wang2019normalized}, Wild6D~\cite{fu2022category}, and Objectron~\cite{ahmadyan2021objectron}) because they are fragile and costly to replace, nor do we include bikes or chairs (Objectron~\cite{ahmadyan2021objectron}) because they are too large.
We also do not include books (from Objectron~\cite{ahmadyan2021objectron} and CO3D~\cite{reizenstein2021iccv:co3d}) because their non-rigidity makes them nearly impossible to grasp with a single robotic hand.
Moreover, we do not include apples, carrots, oranges, bananas, and balls (from CO3D~\cite{reizenstein2021iccv:co3d}) because they do not support functional grasping; although they could be included if only pick-and-place were considered.   

Note that NOCS-REAL275~\cite{wang2019normalized}, PhoCaL~\cite{wang2022phocal}, and HouseCat6D~\cite{jung2022arx:housecat} have a separate scanning process for 3D reconstruction, whereas our dataset and CO3D~\cite{reizenstein2021iccv:co3d} obtain reconstructions from the captures.
Although CO3D~\cite{reizenstein2021iccv:co3d} includes camera poses with respect to each object instance in a single video, there is no correspondence across videos.
For this reason, the poses in CO3D are insufficient to train a category-level object pose estimator.

\section{Method}

In this section we describe the methodology we used for collecting and annotating the dataset.

\begin{figure*}[h]
\centering
\includegraphics[width=0.98\linewidth]{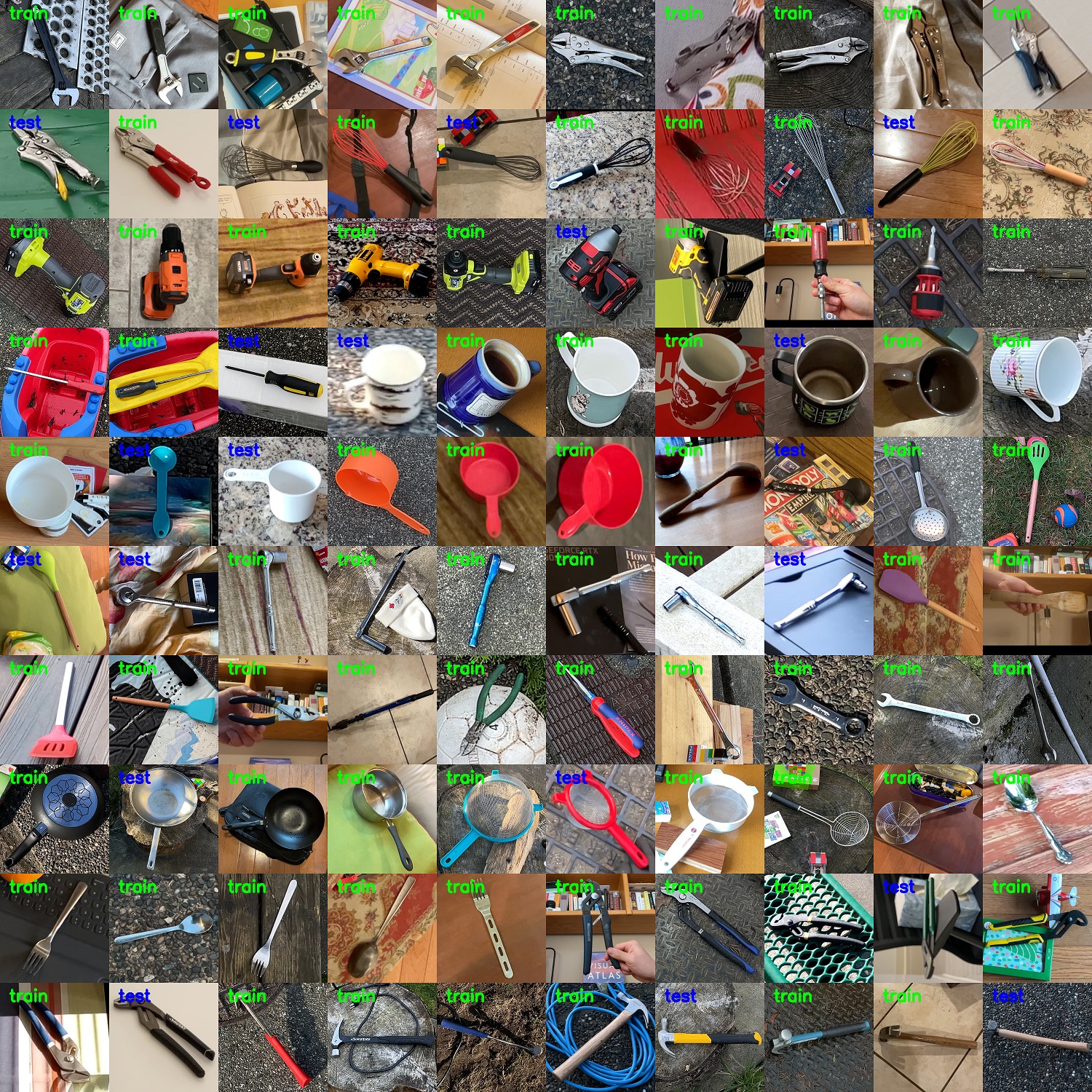}%
\caption{One hundred of the object instances collected in our dataset. The train/test split is shown on the top-left of each thumbnail.}
\label{fig:hundred}
\end{figure*}

\begin{figure}[h]
\centering
\includegraphics[width=0.48\textwidth]{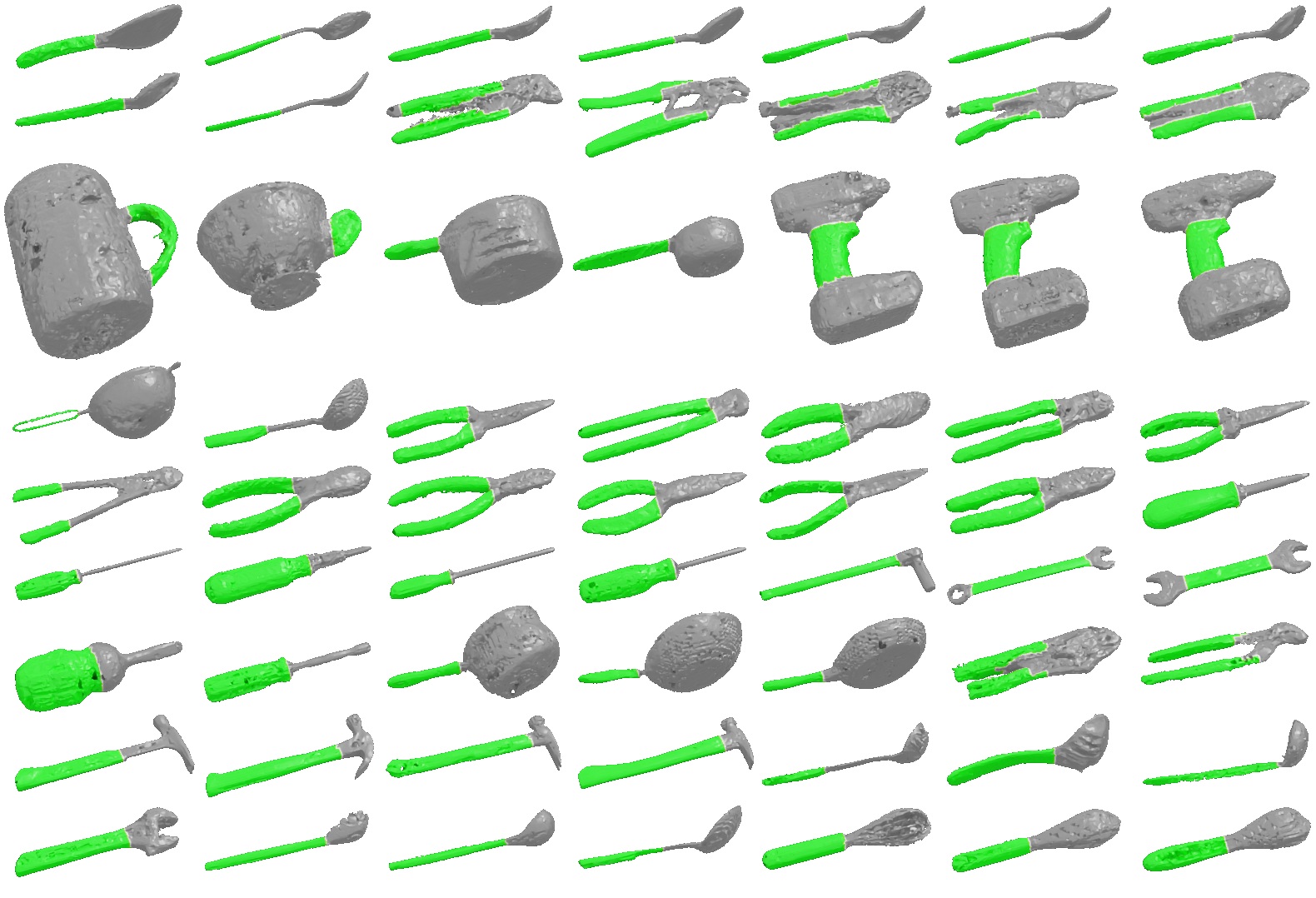}
\caption{Reconstructed meshes with the ground truth affordance annotations showing the handles (green).}
\label{fig:handles}
\end{figure}

\begin{figure*}[h]
\centering
\begin{tabular}{cccc}
    \hspace{-2ex}
    \includegraphics[trim={0 100px 0 125px},clip=true,width=0.24\linewidth]{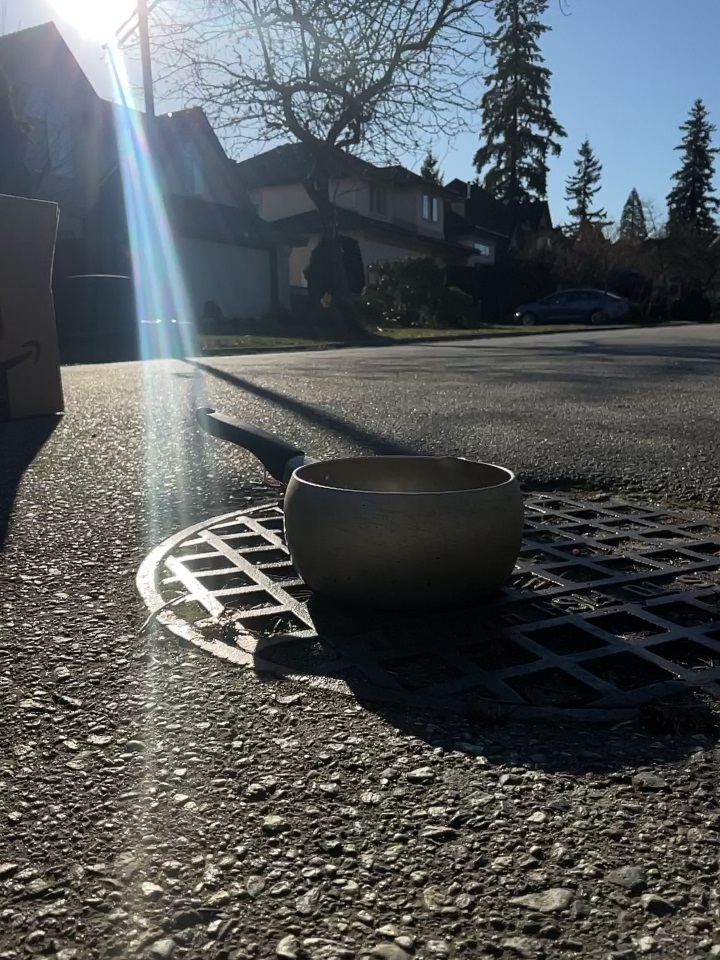} 
    &
    \hspace{-2.5ex}
    \includegraphics[trim={0 125px 0 100px},clip=true,width=0.24\linewidth]{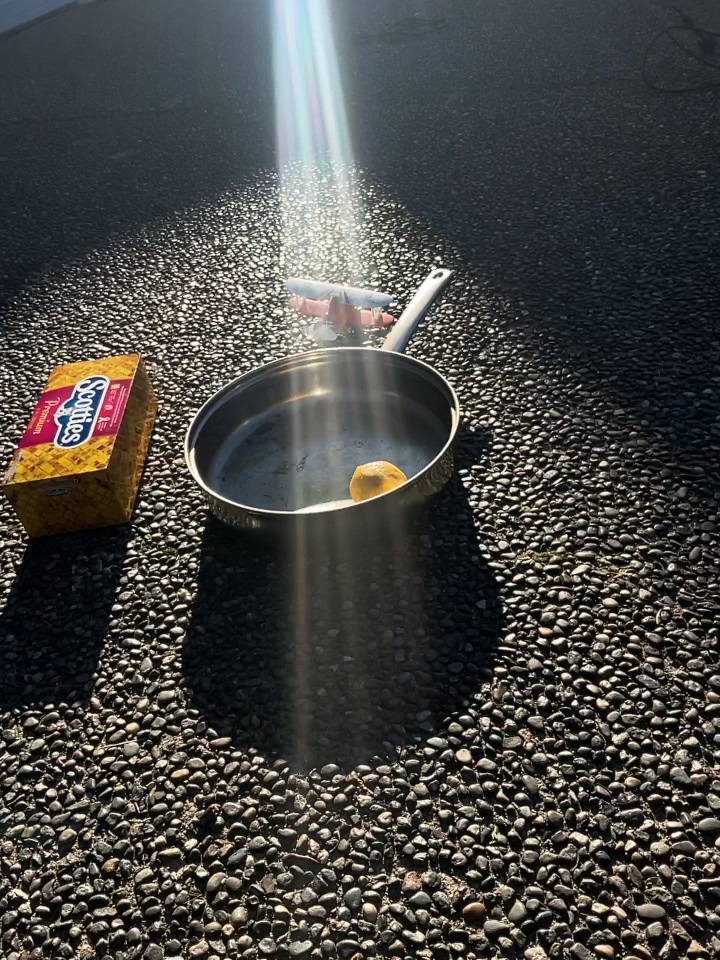} 
    &
    \hspace{-2.5ex}
    \includegraphics[trim={0 125px 0 100px},clip=true,width=0.24\linewidth]{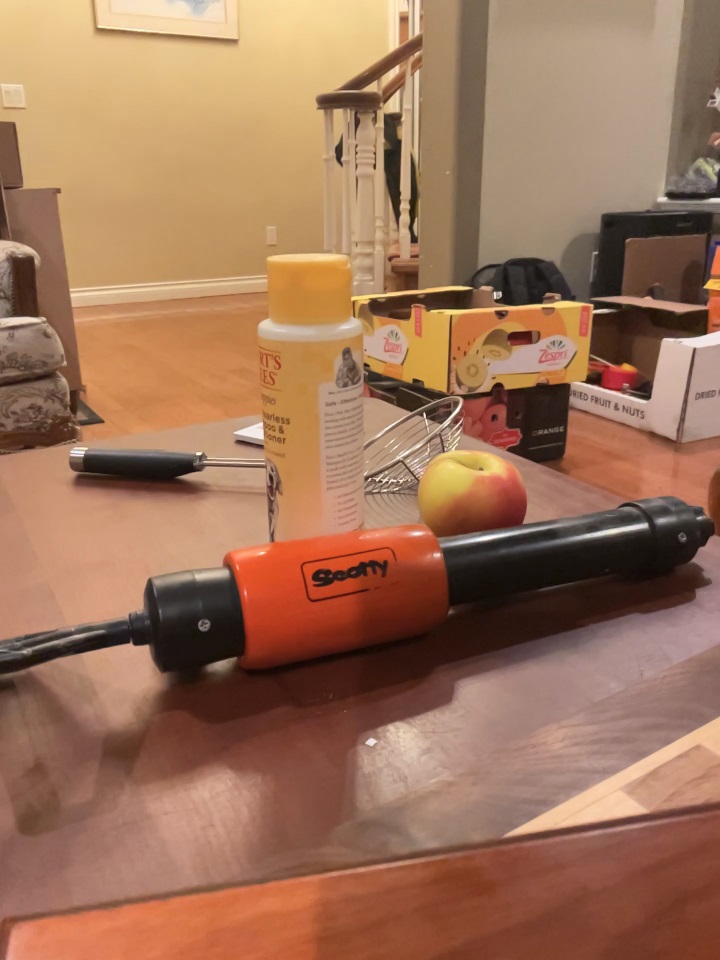} 
    &
    \hspace{-2.5ex} 
    \includegraphics[trim={0 0px 0 20px},clip=true,width=0.24\linewidth]{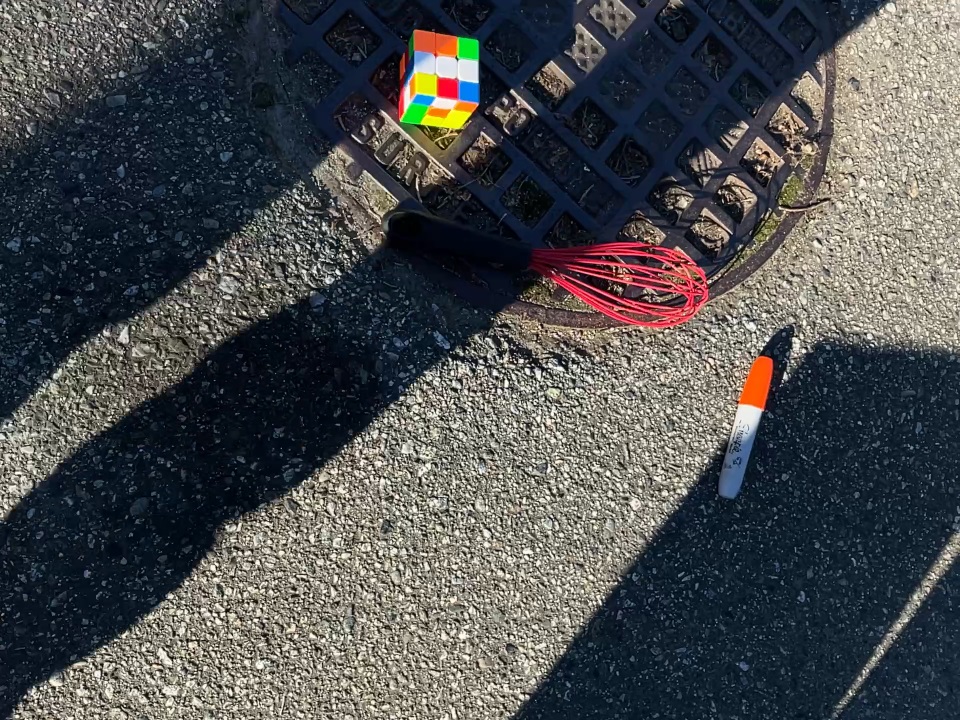} 
    \\
    \hspace{-2ex}
    \includegraphics[trim={0 100px 0 125px},clip=true,width=0.24\linewidth]{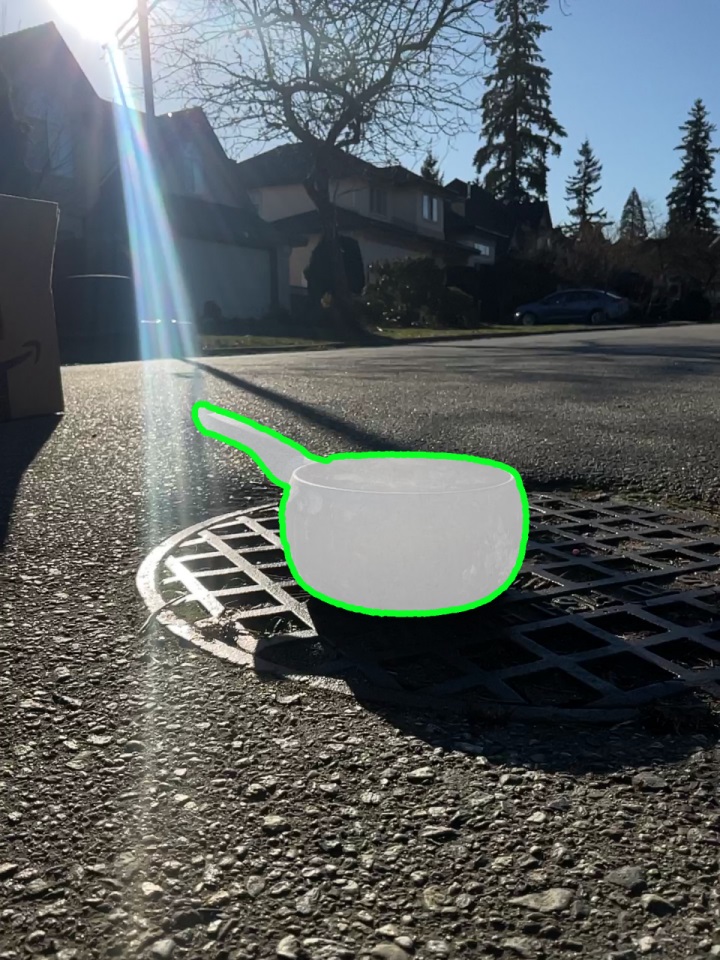} 
    &
    \hspace{-2.5ex}
    \includegraphics[trim={0 125px 0 100px},clip=true,width=0.24\linewidth]{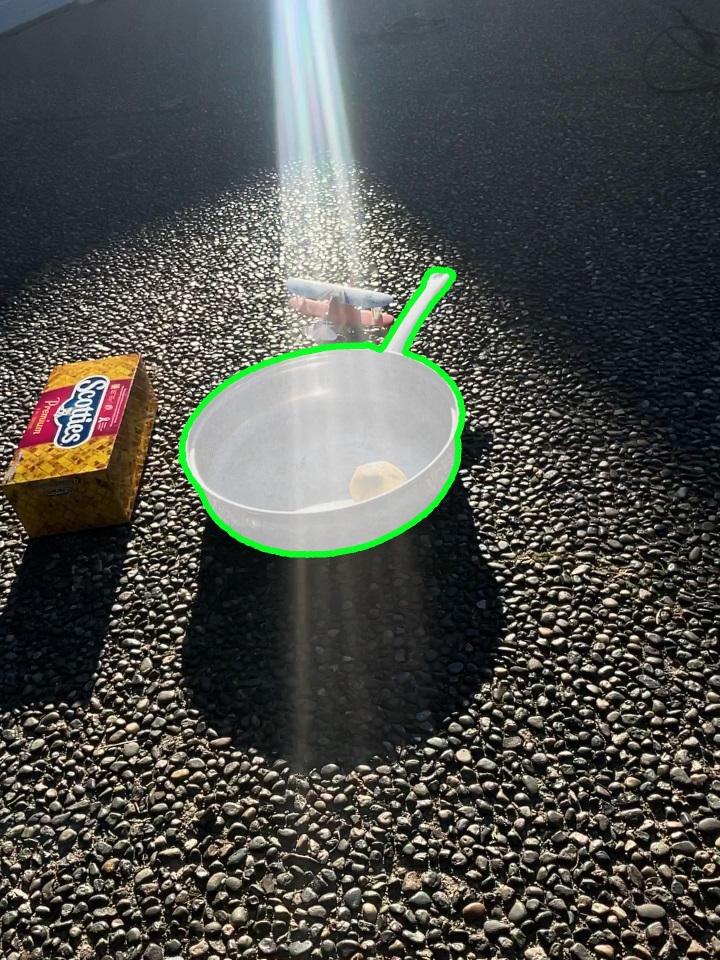}
    &
    \hspace{-2.5ex}
    \includegraphics[trim={0 125px 0 100px},clip=true,width=0.24\linewidth]{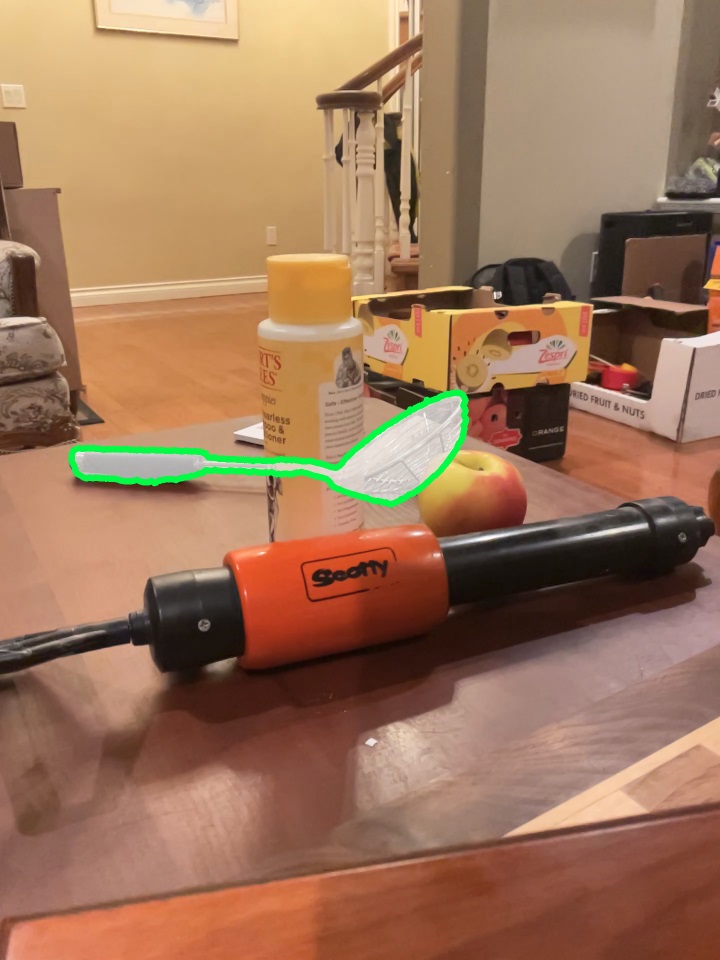}
    &
    \hspace{-2.5ex}
    \includegraphics[trim={0 0px 0 20px},clip=true,width=0.24\linewidth]{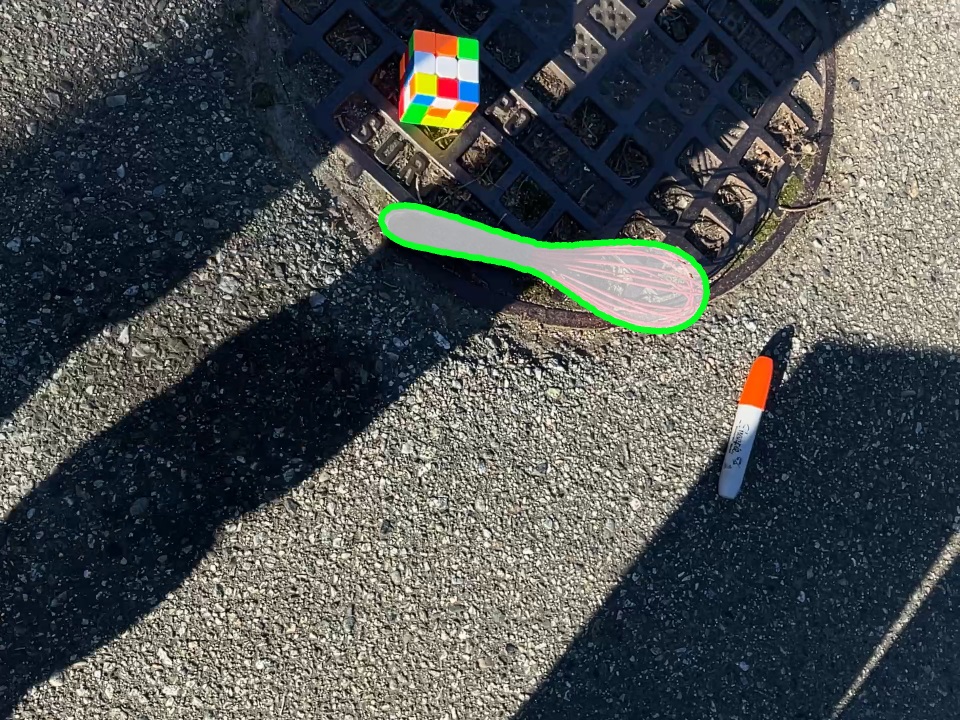}
\end{tabular}
\caption{Challenging images with extreme lighting conditions, glare, shadows, occlusion, reflective surfaces, and perforated objects.  Shown are the original image (top) and object overlay (bottom).
}
\label{fig:glare}
\end{figure*}

\begin{figure*}[h]
\centering
\includegraphics[width=0.98\textwidth]{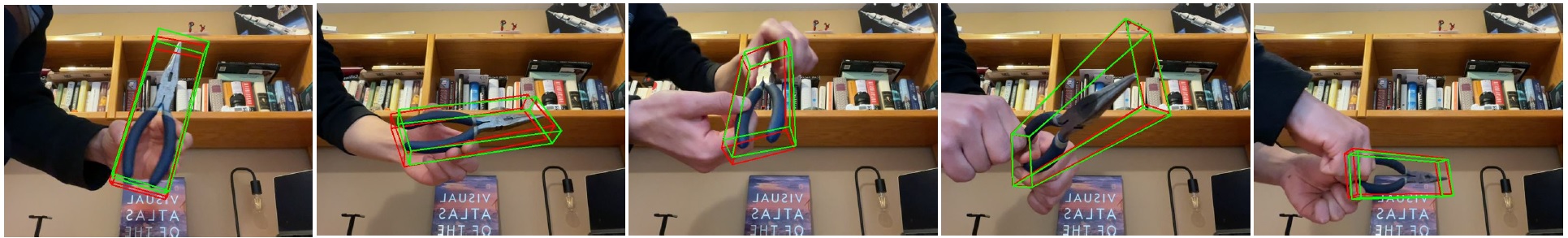}
\caption{Qualitative results of 6-DoF object pose on dynamic pliers sequence (left to right), where a human operator first rotates the object for full reconstruction and then demonstrates the functional affordance. Red box visualizes the raw output by BundleSDF~\cite{wen2023bundlesdf}  and green box visualizes our refined ground-truth.}
\label{fig:qualplier}
\end{figure*}

\subsection{Data collection}

For each category, we acquired at least twelve different object instances, i.e., different brands or sizes, for a total of 212 objects across 17 categories. 
For each object, approximately 10 static scene scans were captured in different lighting conditions, backgrounds, and with different distractor objects. 
Sample images  are shown in Fig.~\ref{fig:hundred}.  

For some objects, an additional reference scan (with the object carefully placed to avoid occlusion and self-occlusion) was captured in order to generate a canonical mesh for registration in the data annotation step. For objects where a reference scan was not captured, we selected the scene mesh with the best reconstructed geometry to be the canonical mesh.  The final mesh of each object was reconstructed using all image frames from all scenes.

Videos were captured at HD~resolution using rear-facing mobile device cameras
while simultaneously estimating rough camera poses by sensor fusion using the built-in ARKit\footnote{\url{https://developer.apple.com/documentation/arkit}} or ARCore\footnote{\url{https://developers.google.com/ar}} libraries. 
Before any other processing, we first selected the sharpest image from each set of $k$ consecutive frames of the video, with $k = \lfloor n_{\text{captured}} / n_{\text{desired}} \rfloor$, where $n_{\text{captured}} \approx 1000$, and $n_{\text{desired}}=120$. This step used the variance of the Laplacian of Gaussian to measure sharpness.

We also captured 51 dynamic scenes with the front-facing TrueDepth camera of the iPad Pro, recording at $640 \times 480$. 
One video was captured per object, using the subset of objects that we found to be amenable to depth sensing.
These captures were divided into two parts.  First, the object was rotated in front of the static camera so as to view it from as many angles as possible to create a quality 3D reconstruction using BundleSDF~\cite{wen2023bundlesdf}. 
Second, the object was grasped by a human in a functional manner (e.g., grabbing the handle) and moved along a functional path in slow motion (e.g., pretending to strike a nail with the hammer). 
By manipulating objects in a natural way like this, we provide somewhat realistic scenarios for object pose estimation correlated with object affordances.
Note that these dynamic captures together with accurate annotations are unique to our dataset, as all previous datasets in this space are restricted to static scenes.

\subsection{Data annotation of static scenes}
Although ARKit/ARCore provide camera pose estimates for every frame, we found these to be too noisy for 3D reconstruction and annotation.\footnote{Specifically, the median difference between AR-provided camera poses and COLMAP-generated poses was 1.8~cm and 11.4 degrees after aligning the transform clouds and accounting for scale.  These errors were large enough to prevent our Instant-NGP process from converging.}
As a result, we instead used COLMAP~\cite{schoenberger2016cvpr:colmap} to estimate unscaled camera poses from the images alone.
We ran COLMAP on the unmodified images (without cropping or segmenting) to take advantage of background texture.
Then we ran XMem~\cite{cheng2022eccv:xmem} to segment the foreground object from the background.  Although this step required a small amount of manual intervention (usually one or two mouse clicks in the first frame), it was otherwise automatic.  
We then ran Instant NGP~\cite{mueller2022instantngp} on the segmented images to create a neural 3D representation of the object, from which we extracted a mesh using Marching Cubes. %
This process was repeated for every video/scene.

Because the camera poses generated by COLMAP are without absolute scale or absolute orientation, we used the captured ARKit/ARCore poses to rescale the COLMAP poses (multiplying by a single scalar) and to align the vertical orientation with respect to gravity. 
The resulting transform was also applied to the 3D reconstructed meshes.

For each category, a canonical coordinate frame convention was established to ensure consistency across instances. 
As a general rule, the axis of symmetry was set as the $x$ axis for categories with rotational symmetry (e.g., screwdrivers) and the plane of symmetry was set as the $x$-$y$ plane for the remaining categories. We aligned all categories such that the ``front" of the object is $+x$ and the ``top" of the object is $+y$. 
For example, hammers were oriented so that the handle is along the $x$ axis, with the head of the hammer at $+x$; the head itself extends along the $y$ axis, with the face being $+y$ and the claw $-y$.

After establishing the coordinate frame of the canonical reference mesh for an object instance, we computed the transform between this reference mesh and each of the meshes for the same object from different scenes. 
Due to noise in the captures that can obscure small details, we were not able to find a tool capable of automatically, reliably, and efficiently calculating this transform.
Instead, we found it more practical to align the videos using a semi-automatic tool that used the extent of the oriented bounding box of the object as an initial alignment, from which the final alignment was obtained interactively. In our experience, it takes no more than a minute to do this alignment, and the alignment is only required once per scene.

Using the transform from the object's COLMAP-coordinate pose to the canonical reference pose, together with the camera poses generated by COLMAP, the pose of the object relative to the camera was computed in every frame.
Occasionally (a few frames in some of the scenes), the pose was incorrect due to errors in the camera poses.
In order to ensure pose quality across the entire dataset, we reprojected the mesh using the computed poses onto the input images and 
manually removed any frames where the overlap of the reprojection was poor.

\subsection{Data annotation of dynamic scenes}

To obtain the ground-truth object segmentation mask for dynamic scenes, we leveraged XMem~\cite{cheng2022eccv:xmem} (as with static scenes), followed by manual verification and correction.
Unlike the static scenes where ground-truth object pose can be trivially inferred from camera pose localization, determining object poses in dynamic scenes is extremely challenging.
To solve this problem, we applied BundleSDF~\cite{wen2023bundlesdf} to these dynamic videos to simultaneously reconstruct the geometry of the object, as well as to track the 6-DoF pose of the object throughout the image frames. Compared to BundleTrack~\cite{wen2021bundletrack}, the additional 3D reconstruction from BundleSDF allows to register to the model created from the static scene so as to unify the pose annotations.
To assess the BundleSDF result, we randomly added noise to the output pose in each frame, then applied ICP to align the mesh with the depth image. 
We then manually inspected all frames to ensure high quality.
For most frames, the output of BundleSDF and ICP are nearly identical, and it is difficult to assess which of the two is more accurate.
Therefore, we obtain ground truth by averaging them.

Because BundleSDF tracks throughout the entire video, it allows for the entire object to be reconstructed---including the underside---which is typically an unsolved problem for static table-top scanning processes. 
Finally we repeated the same global registration step (as in the static scenes) between the obtained mesh to the canonical mesh of the same object instance, in order to compute the transform to the canonical reference frame. 
From this process, we obtained the category-level object pose w.r.t.\ the camera in every frame of the video.

\subsection{Annotating affordances}

For affordances, we manually labeled handle regions in the 3D mesh from each object reconstruction.
Fig.~\ref{fig:handles} shows some 3D object reconstructions with the handles labeled.
Our annotations could be extended to include other affordance-based segmentation or keypoint labels that capture the functional use of the objects.

\section{Results}

\input{tables/stats}

\begin{figure}[h]
\centering
\includegraphics[width=0.48\textwidth]{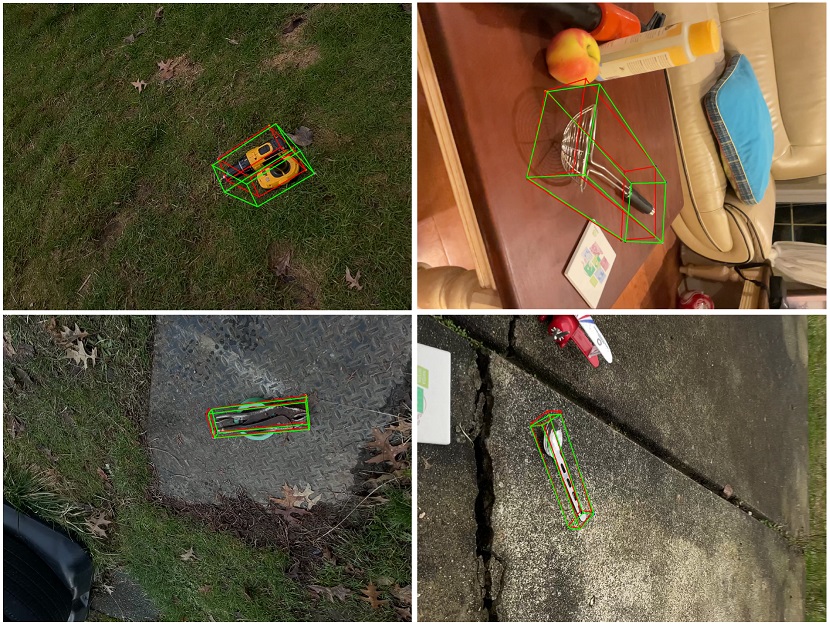} 
\caption{Category-level object detection and 6-DoF pose estimation results from CenterPose~\cite{lin2022icra:centerpose} (red) on test images containing unseen object instances, compared with ground-truth (green).}
\label{fig:centerpose}
\end{figure}

\subsection{Dataset statistics}\label{sec:data_stats}

For each object instance we have a 3D reference mesh, about 10 static RGB captures both with and without clutter, object segmentation masks, 6-DoF pose in all frames with respect to a canonical coordinate frame for the category, and affordance annotations.
Some objects also have a dynamic RGBD video. 
All annotations are stored in standard formats: COCO~\cite{lin2014coco} 2D bounding boxes/instance masks and BOP~\cite{hodan2018eccv:bop} 6-DoF poses.

Our capture protocol was designed to record 360$^\circ$ around each object. 
Such surrounding captures are not common among datasets,
but are important for quality 3D reconstructions, as well as for view diversity.

We ensured that our object selection and captures included extreme lighting conditions, glare, moving shadows, shiny objects, and heavily perforated objects.
See Fig.~\ref{fig:glare}. Within each category, we also purposely chose a diverse range of individual objects. 
While this makes it challenging for categorical pose estimators and object detectors to perform well on our dataset, we expect our dataset to motivate further research in handling these real-world challenges. 
Tab.~\ref{tab:stats} presents the dataset statistics, along with quantitative results of 2D detection and 6-DoF pose estimation, described next.

\subsection{Object detection}\label{sec:detection}

To validate the dataset, we defined a test set by holding out 3 objects from each category (see Fig.~\ref{fig:hundred} for examples), including all scenes in which they appear.
We trained a single Mask-RCNN~\cite{he2017iccv:maskrcnn} model to localize and segment instances from the 17 object categories.
Training was performed with the Detectron2 toolkit\footnote{\url{https://github.com/facebookresearch/detectron2}} using model weights pretrained on the COCO instance segmentation task.
For 2D object detection, we achieved $\text{AP}_{50}\!=\!0.774$ and $\text{AP}_{75}\!=\!0.733$, where $\text{AP}_n$ is average precision at n\% IoU.
Combining thresholds from 50\% to 95\%, $\text{AP}\!=\!0.656$.
For segmentation, we achieved 
$\text{AP}_{50}\!=\!0.776$, $\text{AP}_{75}\!=\!0.657$, and $\text{AP}\!=\!0.562$.
These results suggest that our data set is large and diverse enough to support interesting research in this area, though additional images could help to yield even more robust results.
Our training process could also be augmented with synthetic data, which could be facilitated by our 3D reconstructed models, with some additional work, as discussed below.

\subsection{Category-level object pose estimation}\label{sec:static_pose}

As another proof-of-concept, we show the capability of learning category-level 6-DoF pose estimation using our dataset. Specifically we evaluated the RGB-based category-level pose estimation method CenterPose~\cite{lin2022icra:centerpose} following the same train/test split. Due to the wide diversity of poses, we rotate the image by the increment of 90$^\circ$ that best aligns the vertical axis of the object with the vertical axis of the image, and we assume access to ground-truth object dimensions.  Example qualitative results are demonstrated in Fig.~\ref{fig:centerpose}.

\section{Discussion}

One of our goals in this project has been to explore the democratization of the creation of high-quality 3D datasets for robotic manipulation. This motivation has guided us to use off-the-shelf cameras and to select objects with properties that have been difficult for traditional annotation methods (e.g., reflective utensils, perforated strainers, and thin whisks). 
We aim to make the annotation pipeline as automatic as possible so that  it is realistic for researchers to generate their own datasets.
As a result, we did not outsource any of our annotation or leverage a large team. 
 In fact, once the pipeline was set up, our entire dataset consisting of 308k annotated frames from 2.2k videos was captured and fully processed by a couple researchers over the span of just a couple weeks. 

Overall, we estimate that it takes approximately 5--6 minutes of manual interactive time, on average, to process a single static scene.
This estimate includes both capture time and human intervention of the software process (rewatching the video to verify XMem results, overseeing the coordinate transformation process, and so forth).
The rest of the process is automatic, requiring approximately 20 minutes or so of compute.
Of this time, COLMAP is by far the most compute-intensive part.

Given these time estimates, it is realistic for a single researcher familiar with our pipeline to capture and process a single category, consisting of $\sim$12 objects across $\sim$100 scenes with $\sim$15k annotated frames in diverse settings in about 12 hours---excluding the computation time of COLMAP.
Thus, we believe that our research has made significant advances toward this goal.

While our pipeline has been shown to be scalable, there  are still many bottlenecks requiring manual intervention. 
We had initially created an automatic mesh alignment algorithm using 3D point cloud matching with 3DSmoothNet~\cite{gojcic2019cvpr:3DSmoothNet} but found that it was too slow, taking over 30 minutes for certain objects, and sensitive to scale variations. This led us to adopt a much faster but more manual approach using the oriented 3D bounding box of the mesh. Segmentation using XMem~\cite{cheng2022eccv:xmem} is currently another step in our data annotation pipeline requiring manual intervention. While this is typically minimal for most scenes, requiring no more than a few clicks at the beginning, it still requires 1--2 minutes of supervision. A fully automatic segmentation method would remove this bottleneck and further increase the scalability of our pipeline.

Lastly, meshes exported from Instant NGP currently have poor texture. While these renderings can be improved, baked-in lighting will make it difficult to render the objects realistically in new environments. In addition, there is currently no way to edit the material properties for domain randomization. Overall, these factors preclude generating realistic and high-quality synthetic data for training. 
Fortunately, neural reconstruction and rendering is a rapidly developing field. Further research in this area will add another dimension to our pipeline by allowing it to be used for synthetic data generation in addition to annotation of real data.

\section{Conclusion}

We have presented a large dataset of images annotated with 6-DoF category-level pose and scale for robotics.  
This dataset is the largest non-outsourced of its kind.
As such, it provides lessons on how other researchers can collect such data in the future.
By capturing dynamic scenes, full 360$^\circ$ scans, and occlusions, and providing object affordances and 3D reconstructions, our dataset provides unique characteristics for perceptual learning for robotic manipulation and functional grasping.

\bibliographystyle{IEEEtran}
\bibliography{main}

\end{document}

%% file: tables/comparison.tex
\begin{table*}[h]
    \centering
    \caption{Comparison of category-level object datasets.
    See Section~\ref{sec:compare} for details.
    Only the annotated images of Wild6D and our dataset are considered here.
    $^\dagger$PhoCaL and HouseCat6D also use a polarization camera.
    $^\ddagger$CO3D has an additional 5 categories that can be indiscriminantly grasped but that do not afford functional grasping.
    $^*$Our dataset also includes dynamic videos from an RGBD camera.
    }\label{tab:comparison}
    \begin{tabular}{cccccccccccccccc}
    \toprule
    dataset & modality & cat. & manip. & obj. & vid. & img. & pose & scale 
        & 3D recon.  & stat. & dyn. & 360$^\circ$ & occ. & afford. \\
    \midrule 
    NOCS-REAL275~\cite{wang2019normalized} 
        & RGBD   & \,\,\,6 & \,\,\,4 & \,\,\,42 & \,\,\,18 & \,\,\,\,8k & \greencheckmark & \greencheckmark & \greencheckmark & \greencheckmark & \redxmark & \redxmark & \greencheckmark & \redxmark \\
    PhoCaL~\cite{wang2022phocal} 
        & \,\,\,RGBD$^\dagger$ & \,\,\,8 & \,\,\,8 & \,\,\,60 & \,\,\,24 & \,\,\,\,3k & \greencheckmark & \greencheckmark & \greencheckmark & \greencheckmark & \redxmark & \redxmark & \greencheckmark & \redxmark \\
    HouseCat6D~\cite{jung2022arx:housecat} 
        & \,\,\,RGBD$^\dagger$ & 10 & 10 & 194 & \,\,\,41 & \,\,24k & \greencheckmark & \greencheckmark &  \greencheckmark & \greencheckmark & \redxmark & \greencheckmark & \greencheckmark & \redxmark \\
    Wild6D~\cite{fu2022category} 
        & RGBD & \,\,\,5 & \,\,\,3 &  162 & 486 & \,\,10k & \greencheckmark & \greencheckmark & \redxmark & \greencheckmark & \redxmark & \redxmark & \redxmark & \redxmark \\
    Objectron~\cite{ahmadyan2021objectron}  
        & RGB & \,\,\,9 & \,\,\,4 & \,\,17k & \,\,14k & \,\,\,\,\,4M & \greencheckmark & \greencheckmark & \redxmark & \greencheckmark & \redxmark & \redxmark & \redxmark & \redxmark \\
    CO3D~\cite{reizenstein2021iccv:co3d} 
        &   RGB   & 50 & \,\,\,\,\,\,7$^\ddagger$ & \,\,19k & \,\,19k & 1.5M & \redxmark & \redxmark & \greencheckmark & \greencheckmark & \redxmark & \greencheckmark & \redxmark & \redxmark  \\
    HANDAL (Ours)  
        &  \,\,\,RGB$^*$  & 17 & 17 & 212 & \,\,\,\,\,2k & 308k & \greencheckmark & \greencheckmark & \greencheckmark & \greencheckmark & \greencheckmark & \greencheckmark & \greencheckmark & \greencheckmark  \\   
    \bottomrule
    \end{tabular}
\end{table*}

%% file: tables/stats.tex
\begin{table*}[h]
    \centering
    \caption{Our dataset has 17 object categories.  \emph{Statistics} (left-to-right): the number of object instances, videos, and annotated image frames.  (The latter would increase by about 10$\times$ if non-annotated frames were included.)  \emph{2D metrics}:  average precision (AP) of bounding box detection, and AP of pixel-wise segmentation.  \emph{6-DoF pose metrics}: percentage of frames with 3D intersection over union (IoU) above 50\%; area under the curve (AUC) of percentage of cuboid vertices whose average distance (ADD) is below a threshold ranging from 0 to 10~cm; symmetric version of ADD; and percentage of correct 2D keypoints (PCK) within $ \alpha \sqrt{n}$ pixels, where $n$ is the number of pixels occupied by the object in the image, and $\alpha=0.2$.  For 3D IoU and AUC of ADD / ADD-S, the first number is from RGB only, the second number is from shifting/scaling the final result based on the single ground truth depth value.}\label{tab:stats}

\begin{tabular}{clccccccccc}
  \toprule
  & & \multicolumn{3}{c}{\tH{Dataset Statistics}} & \multicolumn{2}{c}{\tH{2D Detection Metrics}} & \multicolumn{4}{c}{\tH{6-DoF Pose Metrics}} \\
  \cmidrule(lr){3-5} \cmidrule(lr){6-7} \cmidrule(lr){8-11}

  & \tH{Category}      & \tH{Instances} & \tH{Videos} & \tH{Frames} & \tH{BBox AP} & \tH{Seg.~AP} &   \tH{3D IoU} &      \tH{ADD} &    \tH{ADD-S} & \tH{PCK} \\
  \midrule

  & Hammer             &             19 &         195 &       24.0k &        0.757 &        0.624 & 0.552 / 0.812 & 0.568 / 0.710 & 0.581 / 0.721 &    0.772 \\
  & Pliers-Fixed Joint &             12 &         125 &       22.9k &        0.481 &        0.375 & 0.669 / 0.988 & 0.824 / 0.915 & 0.834 / 0.920 &    0.973 \\
  & Pliers-Slip Joint  &             12 &         128 &       17.8k &        0.415 &        0.271 & 0.094 / 0.368 & 0.311 / 0.470 & 0.355 / 0.512 &    0.715 \\
  & Pliers-Locking     &             12 &         130 &       18.0k &        0.767 &        0.674 & 0.146 / 0.430 & 0.344 / 0.481 & 0.385 / 0.541 &    0.717 \\
  & Power Drill        &             12 &         131 &       18.4k &        0.736 &        0.754 & 0.590 / 0.830 & 0.391 / 0.613 & 0.404 / 0.620 &    0.732 \\
  & Ratchet            &             12 &         123 &       15.7k &        0.668 &        0.439 & 0.218 / 0.588 & 0.370 / 0.577 & 0.409 / 0.625 &    0.555 \\
  & Screwdriver        &             12 &         140 &       20.9k &        0.596 &        0.557 & 0.196 / 0.649 & 0.378 / 0.626 & 0.432 / 0.707 &    0.270 \\
  & Wrench-Adjust.     &             13 &         142 &       18.1k &        0.628 &        0.482 & 0.205 / 0.607 & 0.406 / 0.604 & 0.457 / 0.655 &    0.711 \\
  & Wrench-Comb.       &             12 &         133 &       17.6k &        0.790 &        0.523 & 0.256 / 0.599 & 0.534 / 0.688 & 0.568 / 0.730 &    0.767 \\
  \midrule \enspace\multirow{-12}{*}{\sc\begin{sideways}hardware tools\end{sideways}}

  & Ladle              &             12 &         127 &       18.0k &        0.805 &        0.673 & 0.419 / 0.781 & 0.383 / 0.571 & 0.410 / 0.609 &    0.494 \\
  & Measuring Cup      &             12 &         121 &       15.3k &        0.532 &        0.490 & 0.201 / 0.486 & 0.346 / 0.497 & 0.386 / 0.619 &    0.564 \\
  & Mug                &             12 &         130 &       18.3k &        0.532 &        0.551 & 0.668 / 0.905 & 0.503 / 0.667 & 0.507 / 0.722 &    0.695 \\
  & Pot / Pan          &             12 &         127 &       16.1k &        0.796 &        0.807 & 0.629 / 0.795 & 0.353 / 0.531 & 0.373 / 0.552 &    0.634 \\
  & Spatula            &             12 &         131 &       20.2k &        0.456 &        0.395 & 0.155 / 0.532 & 0.211 / 0.450 & 0.251 / 0.505 &    0.408 \\
  & Strainer           &             12 &         127 &       16.0k &        0.656 &        0.624 & 0.235 / 0.681 & 0.239 / 0.487 & 0.260 / 0.536 &    0.247 \\
  & Utensil            &             12 &         118 &       14.8k &        0.740 &        0.566 & 0.160 / 0.477 & 0.368 / 0.521 & 0.405 / 0.589 &    0.606 \\
  & Whisk              &             12 &         125 &       15.9k &        0.791 &        0.751 & 0.427 / 0.786 & 0.370 / 0.562 & 0.443 / 0.641 &    0.431 \\
  \midrule \enspace\multirow{-11}{*}{\sc\begin{sideways}kitchen items\end{sideways}}

  & Total              &            212 &        2253 &      308.1k &        0.656 &        0.562 & 0.340 / 0.670 & 0.407 / 0.588 & 0.440 / 0.637 &    0.605 \\
  \bottomrule
\end{tabular}%
\end{table*}